\documentclass[conference]{IEEEtran}
\IEEEoverridecommandlockouts
\usepackage{cite}
\usepackage{amsmath,amssymb,amsfonts}
\usepackage{algorithmic}
\usepackage{graphicx}
\usepackage{textcomp}
\usepackage{xcolor}
\usepackage{tikz}
\usepackage[table]{xcolor}
\usepackage{tabularx}
\usepackage{array}
\usepackage{comment}
\usepackage{pifont}
\usepackage{booktabs}
\usetikzlibrary{arrows.meta, positioning}

\def\BibTeX{{\rm B\kern-.05em{\sc i\kern-.025em b}\kern-.08em
    T\kern-.1667em\lower.7ex\hbox{E}\kern-.125emX}}
\begin{document}

\title{Bridging the Sim-to-Real Gap under Real-Time Constraints in Autonomous Racing\\
\thanks{This work was supported in part by a donation from AMD to the UCF ECE Department's robotics initiatives.}
}

\author{\IEEEauthorblockN{Hossein Maghsoumi, Yaser P. Fallah}
\IEEEauthorblockA{\textit{Department of Electrical and Computer Engineering} \\
\textit{University of Central Florida}\\
Orlando, USA \\
hossein.maghsoumi@ucf.edu, yaser.fallah@ucf.edu}
}

\maketitle
\IEEEpubid{%
  \begin{minipage}{\textwidth}
    \vspace{6.8\baselineskip}   
    \centering
    \fbox{%
      \parbox{0.92\textwidth}{\centering\small
        This paper has been accepted for presentation at the
        2026 IEEE 104th Vehicular Technology Conference
        (VTC2026-Fall).}%
    }%
  \end{minipage}%
}%
\IEEEpubidadjcol

\begingroup
\renewcommand{\thefootnote}{}
\footnotetext{%
\footnotesize
\textcopyright~2026 IEEE. Personal use of this material is permitted.
Permission from IEEE must be obtained for all other uses, in any current
or future media, including reprinting/republishing this material for
advertising or promotional purposes, creating new collective works, for
resale or redistribution to servers or lists, or reuse of any copyrighted
component of this work in other works.%
}
\addtocounter{footnote}{-1}
\endgroup

\begin{abstract}
Autonomous racing exposes the sim-to-real gap under extreme operating conditions characterized by high speed, tight stability margins, and stringent real-time constraints. 
Although simulation is indispensable for development, controllers that perform well in simulation often degrade abruptly on physical platforms due to interacting effects of dynamics mismatch, estimation delay, and execution-layer latency. This paper frames sim-to-real transfer in autonomous racing as a full-stack, real-time systems problem. We introduce a structured three-layer perspective (Physical/Cyber/Execution) to analyze how mismatches propagate and amplify through closed-loop feedback. We present diagnostic metrics beyond nominal lap time, including performance flip, stability-oriented measures, sensitivity to delay and noise, and latency distribution characterization. 
Mitigation strategies are synthesized from a deployment-oriented viewpoint, emphasizing execution-aware and delay-aware design. Finally, we outline benchmarking guidelines that enable reproducible and fair sim-to-real evaluation under compute and timing constraints. The resulting framework clarifies cross-layer failure mechanisms and provides practical design principles for deployable autonomous racing systems operating near dynamic limits.
\end{abstract}


\section{Introduction}

Autonomous racing has emerged as a high-impact testbed for advancing autonomous driving, precisely because it exposes failure modes that remain latent in conventional urban-driving benchmarks. 
Unlike urban autonomy, which typically operates with conservative safety margins and longer reaction times, racing systems must sustain high-speed operation near tire friction limits, where small modeling errors and millisecond-level delays can trigger instability or loss of control~\cite{9790832, 11077481}. 
Consequently, autonomous racing constitutes a worst-case setting for \emph{sim-to-real} transfer: successes in simulation often fail to reproduce on physical platforms when timing, sensing, and dynamics are stressed simultaneously.

Simulation remains indispensable for development and evaluation due to its safety, scalability, and repeatability. 
However, the sim-to-real gap is particularly severe in racing because discrepancies do not merely reduce performance gradually; rather, they can cause abrupt transitions from stable tracking to oscillation, saturation, or complete loss of control. 
While prior work has explored improved modeling, learning-based generalization, and robust control, many deployments still under-report or under-model real-time execution properties such as pipeline latency, jitter, and control-loop scheduling, despite their outsized impact at high speed.

This paper argues that sim-to-real transfer in autonomous racing should be treated as a \emph{full-stack, real-time} problem: physical mismatches (e.g., tires and friction), cyber-level sensing/estimation imperfections (e.g., sparsity and localization delay), and execution constraints (e.g., compute budgets and scheduling jitter) jointly determine deployability. 
We provide a concise framework to (i) structure these gap sources, (ii) measure their practical impact, and (iii) connect mitigation choices to system-level constraints.

\subsection{Contributions}
The main contributions of this paper are:
\begin{itemize}
    \item \textbf{A racing-specific taxonomy of sim-to-real gaps} that explicitly separates \emph{Physical} (dynamics/environment), \emph{Cyber} (sensing/estimation), and \emph{Execution} (compute/scheduling/latency) factors, emphasizing how they interact under tight stability margins.
    \item \textbf{A measurement and diagnosis toolkit} for sim-to-real degradation in racing, including (i) \emph{performance flip} (simulation ranking $\neq$ real ranking), (ii) stability/safety-oriented metrics beyond lap time, and (iii) sensitivity curves to delay/noise and latency distribution reporting.
    \item \textbf{A deployability-oriented synthesis of mitigation methods} that links system identification, domain randomization, delay-aware design, and architectural choices to real-time feasibility and full-stack error propagation.
    \item \textbf{Benchmarking and reporting recommendations} tailored to reproducible sim-to-real claims in racing (hardware specifications, loop rates, and latency histograms), enabling fair comparison under compute and timing constraints. 
\end{itemize}

The remainder of the paper is organized as follows. 
Section~II summarizes the sim-to-real gap through a three-layer (Physical/Cyber/Execution) perspective and highlights why racing magnifies mismatches. 
Section~III introduces metrics and protocols for measuring the gap under noise and delay. 
Section~IV reviews mitigation strategies from a full-stack, real-time viewpoint and distills design principles for deployable racing stacks. 
Finally, Section~V discusses benchmarking and reproducibility considerations under real-time constraints.

\section{Sim-to-Real Under Real-Time Constraints: A Three-Layer View}

\begin{figure}[t]
\centerline{\includegraphics[width=0.6\linewidth]{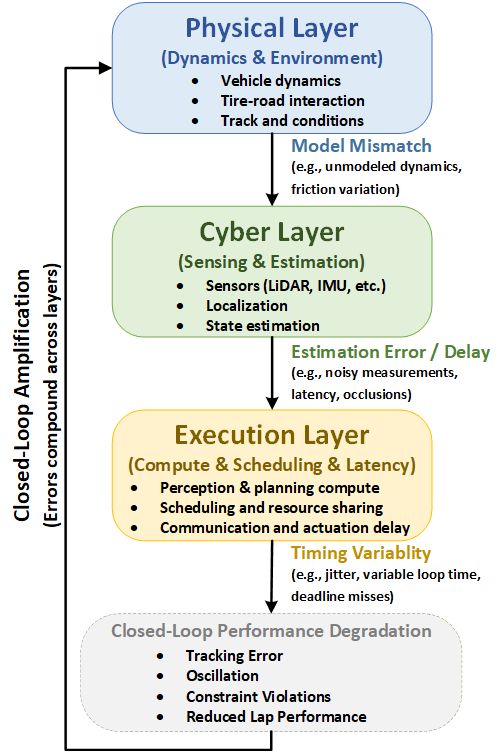}}
\caption{Layered and closed-loop view of sim-to-real gaps in autonomous racing. 
Physical mismatch, estimation delay, and execution-layer latency propagate across the autonomy stack, while control actions feed back into vehicle dynamics, amplifying instability under high-speed operation.}
\label{fig:taxonomy_three_layer}
\end{figure}

Fig.~\ref{fig:taxonomy_three_layer} summarizes the sim-to-real gap in autonomous racing using a three-layer perspective: \emph{Physical}, \emph{Cyber}, and \emph{Execution}. 
Unlike conventional urban-driving scenarios, racing operates near stability limits, where mismatches across these layers interact nonlinearly and are amplified by high speeds and tight timing constraints.

\subsection{Physical Layer: Dynamics and Environment}

The physical layer captures discrepancies between simulated and real vehicle-environment interaction. 
In racing, simplified tire models, constant friction assumptions, and idealized actuator dynamics commonly used in simulation fail to represent nonlinear load transfer, slip behavior, and surface variability \cite{o2020f1tenth}, \cite{chisari2021learning}. 
Small modeling errors that are negligible in urban driving can cause large trajectory deviations when operating near friction limits.

Environmental mismatch further exacerbates this effect. 
Track surface heterogeneity, lighting variation, debris, and temperature-dependent grip alter vehicle behavior in ways rarely captured by static simulators. 
Consequently, control policies or planners tuned in simulation may exploit unrealistically stable dynamics that do not exist on physical hardware.

\subsection{Cyber Layer: Sensing and Estimation}

The cyber layer represents sensing, perception, and state estimation. 
Simulators frequently assume idealized sensing with deterministic timing and simplified noise models. 
In practice, LiDAR sparsity, motion blur, asynchronous sensor streams, and localization drift introduce both error and delay~\cite{9790832}. 

Localization delay is particularly critical in racing. 
At high speed, even small pose latency translates into significant spatial prediction error. 
Planners may generate aggressive trajectories based on outdated states, and controllers tracking these trajectories operate with minimal margin for correction. 
Thus, estimation inaccuracies propagate rapidly downstream.

\subsection{Execution Layer: Compute, Scheduling, and Latency}

The execution layer encompasses compute limitations, middleware overhead, scheduling jitter, and control-loop timing. 
Simulation environments often neglect execution constraints, implicitly assuming synchronous updates and fixed computation times. 
On real platforms, perception, planning, and control share limited compute resources, and latency varies across cycles.

In racing systems, worst-case latency—not average latency—often determines stability. 
Rare scheduling spikes or dropped frames can destabilize a controller operating near dynamic limits. 
Therefore, execution timing must be treated as a first-class system parameter rather than an implementation detail.

\subsection{Cross-Layer Amplification in Racing}

A key insight is that sim-to-real failures rarely originate from a single layer. 
Physical mismatch, estimation delay, and execution jitter interact through closed-loop feedback. 
Under aggressive racing conditions, these interactions can produce cascading instability even when each individual layer appears acceptable in isolation.

This cross-layer amplification explains why policies that perform well in simulation may exhibit abrupt degradation on hardware, a phenomenon we later formalize as \emph{performance flip}. 
Understanding these layered interactions is essential for designing mitigation strategies that remain stable under real-time constraints.

\begin{table*}[t]
\centering
\caption{Metrics for measuring and diagnosing sim-to-real degradation in autonomous racing.}
\label{tab:metrics}
\renewcommand{\arraystretch}{1.15}
\scriptsize
\resizebox{\textwidth}{!}{
\begin{tabular}{lccc}
\hline
\textbf{Metric} &
\textbf{What It Captures} &
\textbf{Sensitive To} &
\textbf{Primary Layer} \\
\hline

Lap Time / Speed
&
Overall task-level performance
&
Dynamics, Estimation, Planning, Timing
&
Cross-Layer
\\

Performance Flip
&
Ranking inconsistency between simulation and hardware
&
Overfitting, Modeling Fidelity, Execution Differences
&
Cross-Layer
\\

Tracking Error
&
Path-tracking accuracy and closed-loop deviation
&
Dynamics Mismatch, Estimation Error, Control Delay
&
Physical / Cyber / Execution
\\

Control Smoothness
&
Steering/throttle variation and actuator demand
&
Latency, Control Frequency, Actuator Dynamics
&
Physical / Execution
\\

Constraint Violations
&
Breach of stability or safety boundaries
&
Estimation Error, Dynamics, Timing
&
Cross-Layer
\\

Latency Histogram (Tail Latency)
&
Execution-time distribution and tail-latency behavior
&
Scheduling Jitter, Compute Load
&
Execution
\\

Noise Sensitivity Curves
&
Robustness to sensing perturbations
&
Localization Noise, Sensor Sparsity
&
Cyber
\\

Delay Sensitivity Curves
&
Stability degradation under timing perturbations
&
Sensing, Estimation, and Control-Loop Delay
&
Cyber / Execution
\\

\hline
\end{tabular}
}
\end{table*}

\section{Measuring and Diagnosing the Sim-to-Real Gap}

Qualitative discussion of sim-to-real failures is insufficient for engineering progress. 
In autonomous racing, degradation often manifests abruptly rather than gradually, necessitating metrics that capture stability boundaries, timing sensitivity, and ranking inconsistency between simulation and hardware. 
Table~\ref{tab:metrics} summarizes evaluation dimensions particularly relevant to racing systems. As shown in Table~I, different metrics expose vulnerabilities in distinct sim-to-real layers, enabling structured diagnosis rather than isolated performance reporting.

\subsection{Beyond Lap Time: Stability-Oriented Metrics}

Lap time and average speed remain intuitive indicators of performance; however, they do not reveal proximity to instability. 
Racing controllers frequently operate near friction and actuation limits, where small perturbations can trigger oscillations or constraint violations. 
Tracking error bounds, steering-rate variance, and cumulative control effort provide more diagnostic insight into robustness under dynamic stress.

Constraint violations (e.g., track boundary excursions or actuator saturation) are particularly informative, as they indicate failure of stability margins rather than mere suboptimality. 
Together, these metrics provide diagnostic insight into sim-to-real transferability beyond nominal performance alone.
\subsection{Performance Flip}

A recurring phenomenon in racing platforms is \emph{performance flip}: a policy that ranks highest in simulation performs worse than alternatives on real hardware. 
This typically arises when simulation artifacts (e.g., unrealistically smooth dynamics or idealized sensing) are implicitly exploited during optimization.

Performance flip highlights the limitation of simulation-only ranking. 
Comparative evaluation across multiple controllers, both in simulation and on hardware, is therefore essential to assess transfer validity \cite{ComparingDRLRacing2023}.

\subsection{Sensitivity to Noise and Delay}

\begin{figure}[h]
\centerline{\includegraphics[width=0.9\linewidth]{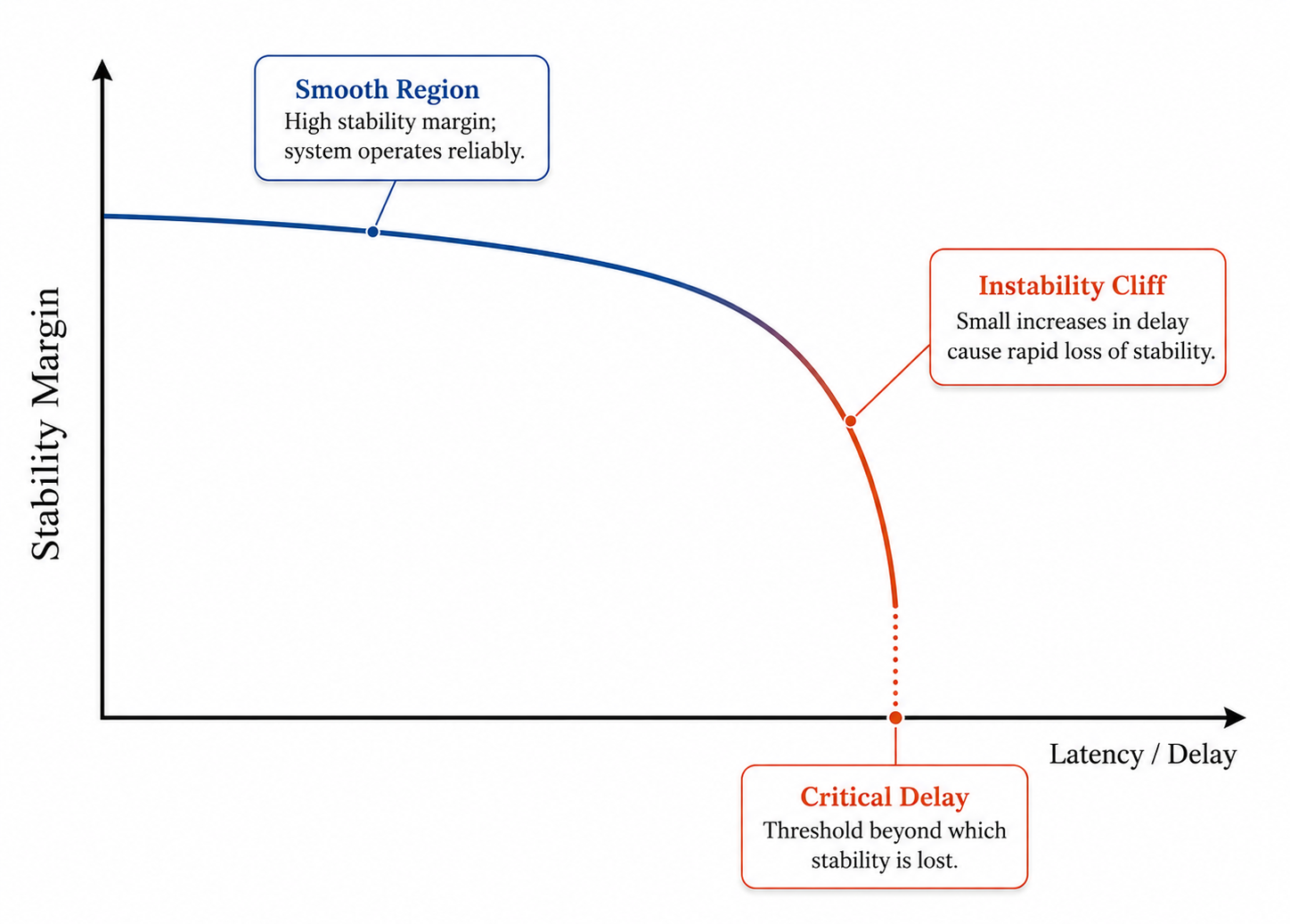}}
\caption{Conceptual illustration of stability degradation with increasing latency. 
Autonomous racing systems often exhibit sharp performance cliffs rather than gradual degradation, emphasizing the importance of tail-latency characterization.}
\label{fig:latency_cliff}
\end{figure}

Simulation provides the unique ability to inject controlled perturbations. 
Systematic variation of localization noise, sensing delay, and control-loop frequency reveals sensitivity boundaries. 
Steep degradation curves with respect to delay or noise indicate fragile designs unlikely to transfer robustly to hardware. This cliff-like degradation behavior is conceptually illustrated in Fig.~\ref{fig:latency_cliff}.

In racing contexts, delay sensitivity is particularly critical. 
At high speeds, even small additional latency translates into substantial spatial prediction error, amplifying downstream control instability.

\subsection{Latency Distributions Rather Than Averages}

Average latency alone is an inadequate descriptor of real-time behavior. 
Racing stability is often governed by worst-case or tail latency events rather than mean execution time. 
Latency histograms or percentile metrics provide a more accurate representation of execution-layer risk.

Reporting end-to-end latency distributions across perception, planning, and control stages enables reproducible comparison across platforms with different compute budgets and middleware configurations. 
Without such reporting, it is difficult to disentangle algorithmic advances from hidden execution advantages.

\subsection{Diagnostic Value of Combined Metrics}

Sim-to-real degradation cannot be adequately characterized using a single metric. 
Lap time, stability margins, control smoothness, and latency statistics should therefore be interpreted jointly. 
For instance, a controller may achieve a competitive lap time while operating with aggressive control actions and heavy tail latency, indicating limited robustness despite strong nominal performance.
Combining these metrics enables systematic diagnosis and supports targeted mitigation strategies.

Table~\ref{tab:literature} compares representative autonomous racing studies across dynamics, sensing and estimation, execution, quantitative sim-to-real evaluation, and full-stack analysis. The comparison shows that real-hardware validation is common, whereas detailed full-stack and real-time characterization remains uneven.
These observations motivate mitigation strategies that explicitly account for cross-layer interactions.

\newcommand{\cmark}{\ding{51}}
\newcommand{\pmark}{$\blacktriangle$}
\newcommand{\xmark}{\ding{55}}

\begin{table*}[t]
\centering
\caption{Representative autonomous racing studies and the sim-to-real and real-time dimensions they address.}
\label{tab:literature}
\renewcommand{\arraystretch}{1.15}
\resizebox{\textwidth}{!}{
\begin{tabular}{lcccccccc}
\toprule
\textbf{Work} &
\textbf{Platform} &
\textbf{Primary Focus} &
\textbf{Dynamics / Model Gap} &
\textbf{Sensing / Estimation} &
\textbf{Execution / RT} &
\textbf{Quantitative Sim-to-Real} &
\textbf{Full-Stack Analysis} &
\textbf{Real Hardware} \\
\midrule

F1TENTH~\cite{o2020f1tenth}
&
F1TENTH
&
Platform and benchmarks
&
\cmark
&
\cmark
&
\pmark
&
\cmark
&
\pmark
&
\cmark
\\

Learning from Simulation~\cite{chisari2021learning}
&
1:43 Kyosho
&
Model randomization and policy regularization
&
\cmark
&
\pmark
&
\pmark
&
\cmark
&
\xmark
&
\cmark
\\

Latent Imagination~\cite{9811650}
&
F1TENTH
&
Model-based RL and zero-shot transfer
&
\pmark
&
\cmark
&
\pmark
&
\cmark
&
\xmark
&
\cmark
\\

Comparing DRL Architectures~\cite{ComparingDRLRacing2023}
&
F1TENTH
&
Architecture-dependent transfer
&
\pmark
&
\cmark
&
\cmark
&
\cmark
&
\pmark
&
\cmark
\\

Train in Austria, Race in Montecarlo~\cite{bosello2022train}
&
F1TENTH
&
LiDAR-based DQN transfer and generalization
&
\pmark
&
\cmark
&
\cmark
&
\pmark
&
\xmark
&
\cmark
\\

ForzaETH Race Stack~\cite{baumann2024forzaeth}
&
F1TENTH
&
Integrated real-time racing stack
&
\pmark
&
\cmark
&
\cmark
&
\pmark
&
\cmark
&
\cmark
\\

\bottomrule
\end{tabular}
}

\vspace{0.4em}
\footnotesize{
\textbf{Legend:}
\cmark~explicitly addressed or evaluated;\;
\pmark~partially addressed;\;
\xmark~not addressed or evaluated.
RT = real-time.
}
\end{table*}

\section{Mitigation and Full-Stack Design Under Real-Time Constraints}

Mitigating the sim-to-real gap in autonomous racing requires more than improving individual models or controllers. 
As discussed in Sections II and III, failures often arise from cross-layer interactions amplified by real-time constraints. 
This section synthesizes mitigation strategies through a deployability-oriented, full-stack lens.

Table~\ref{tab:mitigation} maps representative mitigation and evaluation strategies to sim-to-real layers and highlights their limitations. The table indicates that no single strategy fully
covers all layers, reinforcing the need for coordinated full-stack co-design.

\subsection{Improving Physical Fidelity: Modeling and Identification}

Enhancing simulator fidelity through system identification reduces gross dynamics mismatch. 
Estimating parameters such as tire stiffness, friction coefficients, mass distribution, and actuator delay improves alignment between simulated and real behavior. 
However, racing dynamics vary with speed, temperature, and surface condition; thus, static parameter calibration is insufficient \cite{o2020f1tenth}.

Adaptive or online identification can track parameter drift but introduces computational overhead and potential estimation latency. 
Under tight real-time budgets, the benefit of improved modeling must be balanced against execution feasibility.

\subsection{Robustification Through Randomization and Curriculum}

Domain randomization exposes controllers to distributions of dynamics, noise, and environmental variation during training. 
When properly bounded, it improves robustness to moderate mismatch. 
However, overly aggressive randomization may lead to conservative policies that underperform in racing scenarios.

Curriculum strategies—gradually increasing speed, reducing sensing fidelity, or injecting delay—help controllers develop robustness incrementally. 
Nevertheless, these approaches primarily address modeling and sensing uncertainty; they do not eliminate execution-layer timing issues.

\subsection{Delay-Aware and Execution-Aware Design}

Given the sensitivity of racing systems to latency, mitigation must explicitly incorporate timing into design. 
Delay-aware control techniques integrate sensing and computation delay into prediction or state augmentation frameworks. 
Similarly, reducing architectural depth and avoiding unnecessary synchronization barriers can lower worst-case latency \cite{baumann2024forzaeth}.

Execution-aware evaluation is equally important. 
Algorithms should be validated under realistic compute budgets and scheduling conditions rather than idealized simulation timing. 
Controllers that remain stable under injected delay and reduced loop frequency demonstrate stronger real-world transfer potential.

\subsection{Architectural Trade-Offs and Graceful Degradation}

Autonomy architectures differ in how they balance performance, interpretability, and timing robustness. 
Localization-heavy pipelines may achieve high precision in simulation but become fragile under drift or delay. 
Conversely, reactive or hybrid architectures that combine global planning with local corrective control can degrade more gracefully under mismatch.

A key design principle for racing stacks is \emph{graceful degradation}: performance should decline smoothly as noise or latency increases, rather than exhibiting abrupt failure. 
Design choices that favor moderate aggressiveness, bounded control effort, and predictable timing behavior typically transfer more reliably.

\subsection{From Algorithms to Systems Engineering}

Ultimately, sim-to-real mitigation in racing is a systems engineering problem. 
Algorithmic sophistication alone does not guarantee deployability. 
Successful stacks co-design modeling fidelity, estimation robustness, controller structure, and real-time execution within fixed compute constraints.

By evaluating mitigation strategies through the combined lenses of physical fidelity, cyber robustness, and execution timing, designers can better anticipate cross-layer amplification and reduce the likelihood of performance flip during deployment.

\begin{table*}[t]
\centering
\caption{Mapping of mitigation and evaluation strategies to sim-to-real layers and their limitations.}
\label{tab:mitigation}
\renewcommand{\arraystretch}{1.15}
\scriptsize

\resizebox{\textwidth}{!}{
\begin{tabular}{lcccl}
\hline
\textbf{Strategy} &
\textbf{Physical Layer} &
\textbf{Cyber Layer} &
\textbf{Execution Layer} &
\textbf{Key Limitation} \\
\hline

System Identification
&
\cmark
&
\xmark
&
\xmark
&
Static or nominal calibration may not track changing operating conditions
\\

Domain Randomization
&
\cmark
&
\cmark
&
\xmark
&
Performance depends on distribution coverage and may become conservative
\\

Curriculum Learning
&
\pmark
&
\pmark
&
\pmark
&
Does not guarantee robustness beyond the scheduled perturbations
\\

Delay-Aware Control
&
\xmark
&
\cmark
&
\cmark
&
Requires accurate delay models or bounds and may increase computation
\\

Execution-Aware Evaluation
&
\xmark
&
\xmark
&
\cmark
&
Diagnoses execution-induced degradation but does not directly mitigate it
\\

Hybrid / Residual Architectures
&
\cmark
&
\pmark
&
\pmark
&
Increased integration, tuning, and verification complexity
\\

\hline
\end{tabular}
}

\vspace{0.35em}
\footnotesize{
\textbf{Legend:}
\cmark~directly addresses;\;
\pmark~partially addresses;\;
\xmark~not primarily targeted.
}
\end{table*}

\section{Benchmarking and Reproducible Evaluation}

Reliable progress in sim-to-real autonomous racing requires standardized benchmarking and transparent system-level reporting. 
Without consistent evaluation practices, it is difficult to distinguish genuine algorithmic improvements from hidden advantages in hardware, compute resources, or middleware configuration.

\subsection{System-Level Reporting Requirements}

To support fair comparison, experimental reports should include explicit disclosure of:
\begin{itemize}
    \item Onboard hardware specifications (CPU/GPU, memory, sensors, actuation limits),
    \item Control loop frequencies for perception, planning, and control,
    \item Middleware architecture and synchronization assumptions,
    \item Compute utilization and timing characteristics.
\end{itemize}

Such reporting enables contextual interpretation of performance results and prevents conflating algorithmic advances with execution-layer differences.

\subsection{Latency Characterization and Tail Behavior}

Given the sensitivity of racing systems to timing, average latency is insufficient as a performance descriptor. 
Worst-case and tail latency events often dominate stability outcomes. 
Latency histograms or percentile-based metrics (e.g., 95th/99th percentile) across the autonomy pipeline provide a more realistic assessment of execution robustness.

Reporting end-to-end latency distributions allows meaningful comparison between platforms operating under different compute budgets and scheduling policies. 
This is particularly important when evaluating controllers near dynamic limits.

\subsection{Stress-Test Scenarios for Robustness Evaluation}

In addition to nominal performance evaluation, standardized stress tests improve sim-to-real diagnosis. 
These may include:
\begin{itemize}
    \item Injected sensing delay or localization noise,
    \item Reduced control-loop frequency,
    \item Perturbed friction or dynamics parameters,
    \item Partial sensing dropout.
\end{itemize}

Sensitivity curves derived from such perturbations reveal stability margins and expose fragile designs that would otherwise appear competitive in nominal simulation.

\subsection{Toward Community Benchmarks}

Small-scale racing platforms provide a promising basis for reproducible sim-to-real benchmarking due to their standardized hardware and open-source stacks. 
However, consistent benchmark tasks, reporting templates, and stress-test protocols are still evolving.

Establishing community-aligned evaluation guidelines—centered on full-stack behavior and real-time constraints—would significantly strengthen the credibility and comparability of sim-to-real claims in autonomous racing.

\section{Conclusion}

Autonomous racing exposes the sim-to-real gap under stringent dynamic and real-time constraints, where minor mismatch can trigger abrupt instability. 
Operating near friction and timing limits amplifies cross-layer interactions across physical modeling, sensing and estimation, and execution timing.
As a result, simulation success alone is insufficient to guarantee deployability. This paper framed sim-to-real transfer in racing as a full-stack, real-time systems problem. 
We introduced a three-layer perspective (Physical/Cyber/Execution), proposed diagnostic metrics beyond nominal lap time, and synthesized mitigation strategies through the lens of timing-aware and execution-aware design. 
We further emphasized benchmarking practices that account for hardware configuration, control-loop rates, and latency distributions to enable reproducible and fair evaluation.
The central insight is that robust sim-to-real transfer in autonomous racing requires co-design across modeling fidelity, estimation robustness, controller structure, and real-time execution. 
By treating timing and computation as first-class design constraints rather than implementation details, autonomous racing systems can achieve more reliable and predictable real-world performance.

\bibliographystyle{IEEEbib}
\bibliography{refs}

\end{document}